\documentclass[letterpaper, 10 pt, conference]{ieeeconf}  

\IEEEoverridecommandlockouts                              
\usepackage[utf8]{inputenc}
\usepackage[T1]{fontenc}

\usepackage{graphics} 
\usepackage{epsfig} 
\usepackage{mathptmx} 
\usepackage{mathptmx} 
\usepackage{amsmath} 
\usepackage{amssymb}  
\usepackage{algorithm,tabularx}
\usepackage[noend]{algpseudocode}
\usepackage{eqparbox}
	\usepackage{url}
\usepackage{color}

\title{\LARGE \bf
A Method to use Nonlinear Dynamics in a Whisker Sensor for Terrain Identification by Mobile Robots
}

\author{Zhenhua Yu $^{1}$, S.M.Hadi Sadati$^{2}$, Hasitha Wegiriya$^{3}$,  Peter Childs $^{4}$,Thrishantha Nanayakkara$^{5}$, 
\thanks{*This work was  supported in part by the Engineering and Physical Sciences Research
Council (EPSRC) RoboPatient project under Grant EP/T00603X/1, MOTION project under Grant EP/N03211X/2, Grant EP/N029003/1, and supported in part by China Scholarship Council.}
\thanks{$^{1}$Zhenhua Yu is with Dyson School of Design Engineering,
        Imperial College London, SW7 2DB London, UK
        $\bf z.yu18@imperial.ac.uk$}%
\thanks{$^{2}$S.M.H. Sadati is with the Department of Surgical and Interventional
Engineering, King’s College London, London WC2R 2LS, U.K}
\thanks{$^{3}$ H. Wegiriya is with the Faculty of Natural and Mathematical Sciences, King's College London, London, WC2R 2LS, UK .}%

\thanks{$^{4}$ P.Childs is with Dyson School of Design Engineering, Imperial College London, Dyson Building, 25 Exhibition Road, London, SW7 2DB, UK.}%
\thanks{$^{5}$ T. Nanayakkara is with Dyson School of Design Engineering, Imperial College London, Dyson Building, 25 Exhibition Road, London, SW7 2DB, UK.}%
}

\begin{document}

\maketitle
\thispagestyle{empty}
\pagestyle{empty}

\begin{abstract}

This paper shows analytical and experimental evidence of using the vibration dynamics of a compliant whisker for accurate terrain classification during steady state motion of a mobile robot. A Hall effect sensor was used to measure whisker vibrations due to perturbations from the ground. Analytical results predict that the whisker vibrations will have a dominant frequency at the vertical perturbation frequency of the mobile robot sandwiched by two other less dominant but distinct frequency components. These frequency components may come from bifurcation of vibration frequency due to nonlinear interaction dynamics at steady state. Experimental results also exhibit distinct dominant frequency components unique to the speed of the robot and the terrain roughness. This nonlinear dynamic feature is used in a deep multi-layer perceptron neural network to classify terrains. We achieved 85.6\% prediction success rate for seven flat terrain surfaces with different textures. 

\end{abstract}
\IEEEoverridecommandlockouts

\section{INTRODUCTION}

Terrain  surface identification is an important function for mobile robots moving and performing tasks in extreme,unstructured environments \cite{siegwart2011introduction}. By identifying different terrain surfaces, mobile robots could better perceive the surrounding environment information and decide the optimal next move. For example, they can avoid less favorable terrain types such as soft sand, and use the identification information for mapping and localization objects.  Although many sensors such as cameras are used for object recognition, whiskered tactile perception can potentially provide a robust, and economical solution for these problems, especially in extreme (e.g. dark, foggy) environments.

\begin{figure}[t!] 
\centering
\includegraphics[width=3.5in]{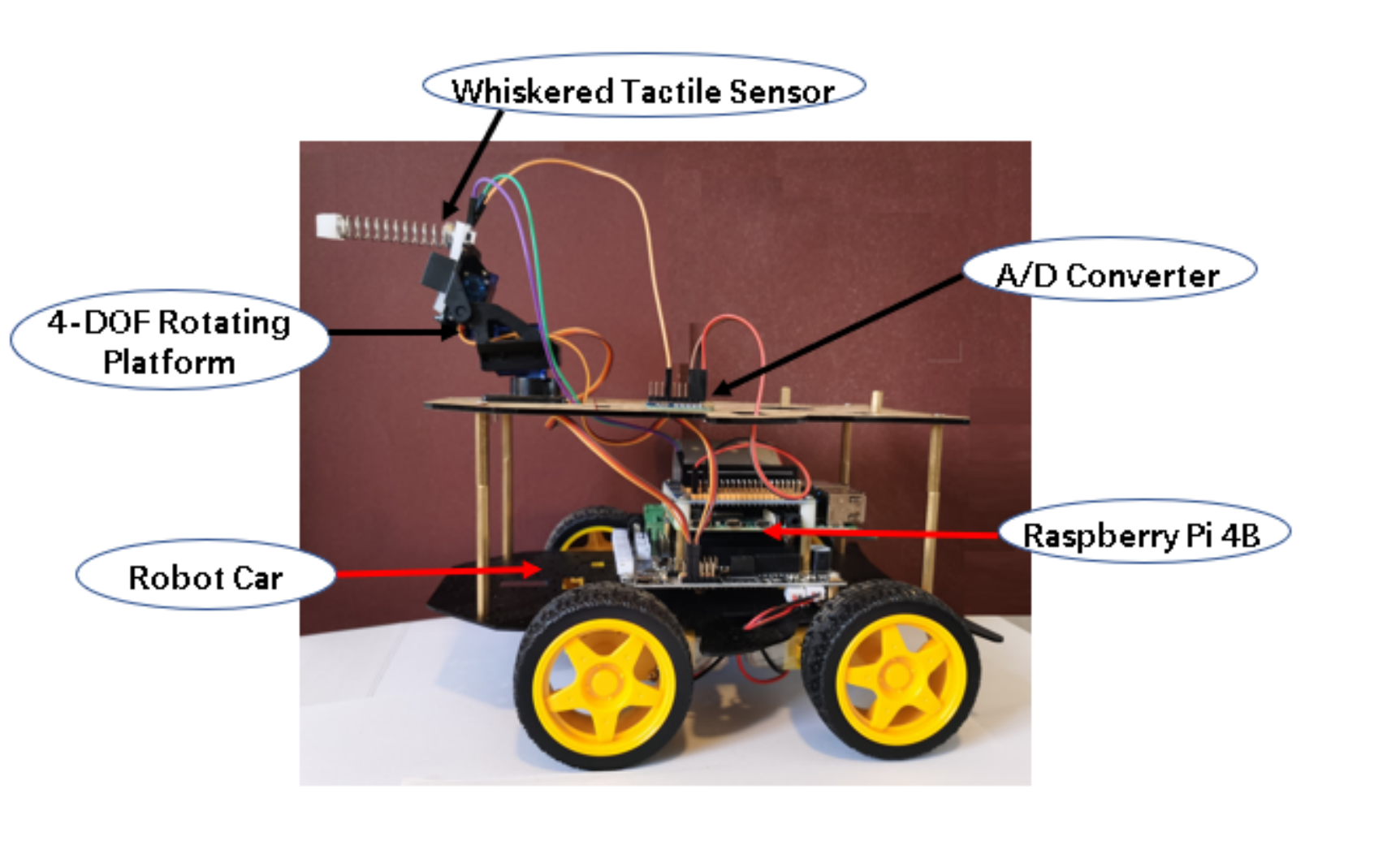}
\caption{Proposed mobile robot equipped with a bioinspired whisker sensor mounted on a 4-DOF servo fusion rotating platform.}
\label{Experiment Car}
\end{figure}

Over the years, vision \cite{howard2001vision}, lidar \cite{suger2015traversability}, sound \cite{valada2017deep}, inertial measurement units  \cite{oliveira2017speed}, and tactile sensors  \cite{fend2003active} have been studied for surface identification. Vision-based methods combined with deep learning have become a research hotspot and significant scientific and technological developments made. However, vision-based methods can have a high probability of failure because of the similar visual appearance of different terrains or objects. Moreover, 
visual accuracy is severely impaired by environmental conditions such as fog, smoke, low light levels, high-brightness and high-temperature.
Recognition and classification methods for different objects based on lidar data have been popularly used in autonomous cars, but its accuracy will drop in sandy applications and dusty scenes.
For an IMU installed on the mobile robot body, a challenge is that they cannot predict the terrain and object types in advance before the robot has contact.
For example, when the robot detects a change in terrain from concrete to soft sand based on inertial signals, the wheels may have already come into contact with the sand. This can lead to dangerous consequences.
Therefore, it is important to find  robust sensing modes for the environment to achieve terrain identification under extreme unstructured environments and the whisker sensor is a choice worth exploring \cite{huet2017tactile}.

Artificial Whiskers sensors have been demonstrated in several studies that are a particularly high-efficiency method for animals such as rats and sea lions which can use them to navigate in the dark and perceive environmental information and features without vision \cite{salman2016advancing}.

This has motivated researchers to design and construct whiskers  for applications such as  navigation \cite{prescott2009whisking}, obstacle  avoidance \cite{fend2004artificial}, object  detection \cite{salman2018whisker}, size measurement \cite {brecht1997functional},  shape recognization \cite{pearson2019active}, and surface information \cite{kaneko1998active}.
For example, in \cite{zurek2014static}, Zurek showed that static antennae can act as locomotory guides which can  compensate for visual methods  to determine the location and distance of obstacles during fast locomotion.
By rotating the whisker against a sequence of contact points of the object and collecting torque information, Solomon and Hartmann put forward a method to obtain an object's 3D contour information \cite{solomon2010extracting}. 
In 2020, authors of \cite{wegiriya2019stiffness} showed that a novel variable stiffness controllable multi-modal whisker sensor can capture different vibration frequencies by controlling whisking speed and the stiffness of the follicle. In all these cases, experimental results prove that the whisker sensors can effectively  compensate for the shortcomings of vision sensors in an extreme environment.

In the work reported here, we designed and constructed a novel whisker tactile sensor to achieve terrain identification for a mobile robot platform (Fig. \ref{Experiment Car}). Reservoir computing \cite{nakajima2020physical,nakajima2015information} is used in the whisker sensor  to map slight changes in the ground perturbations to distinct steady state frequency components. This paper firstly shows that the steady state response of
nonlinear vibration dynamics can be used to classify  terrain textures even on flat terrain.
The remainder of this paper is organized as follows: Section \ref{sysdesign} introduces the design and construction of the whiskers tactile sensor, and the experimental system including the Raspberry robot, whiskered tactile sensor. The whisker sensor vibration characteristics and modal analysis are studied in Section \ref{theoreticalstudy}.
Section \ref{dnn} reports the whisker vibration data collection and feature extraction as well as algorithm framework based on multi-layer perceptron. Section \ref{results} presents the results and analysis of terrain identification experiments .
Finally, Section \ref{conclusions} concludes this paper's contribution and discuss the future work.

\section{System Design \& Preliminary Analysis}
\label{sysdesign}
\subsection{Bioinspired Whisker Sensor Design}

\begin{figure}[t!] 
\centering
\includegraphics[width=3in]{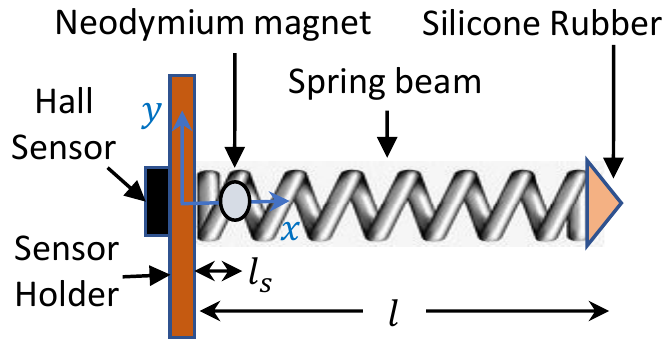}
\caption{The bioinspired whisker sensor.}
\label{Hall Sensor Model}
\end{figure}

The proposed whiskered tactile sensor uses SS49E linear Hall sensors that are orthogonally mounted on one side of a silicone rubber
sensor holder base and spring beam on the other side. The structure is shown in Figure \ref{Hall Sensor Model}.  
A neodymium permanent magnet is embedded inside the spring beam that is made of high-carbon steel, with the free length: 60 mm, wire diameter: 1 mm, outside diameter: 10 mm, and inside diameter: 8 mm respectively. Tapered silicone rubber is installed on the tip of the spring in order to accurately capture the vibration of the surface .
The components and materials cost of this whiskered tactile sensor totalled less than \$5 based on one-off non-discounted prices. The low cost construction of this novel sensor provides an advantage to many existing surface identification sensors

\subsection{Sensor Characterization \& Working Principle}
The whisker sensor is installed on the 4-DOF servo fusion rotating platform  for orientation control of the whisker, which keeps the whisker sensor at a horizontal orientation all the time for clarity.
When the robot traverses different terrain, external vibrations are applied to the whiskered spring shaft, and the shaft deforms also inducing the same vibrations on the magnet inside. This causes a continuous  magnetic flux change near the linear hall effect sensor. Consequently, the hall effect sensor generates continuous low-frequency electrical voltage signals. The  sensitivity of this whiskered tactile sensor relies on the vibration of the whiskered spring beam and the silicone rubber tip.

\subsection{Experimental Setup Design \& Procedure}

The experiment is conducted by a four-wheel mobile robot. The wheeled unit and electronic system are shown in Fig.\ref{Experiment Car}.
The robot is 270 mm in length,  150 mm in height and 130 mm in width,
1.4 kg in mass, with the diameter and width of the wheels are 70 mm and 30 mm respectively. The robot car could traverse through the coarse ground at a speed of up to 1.3 m/s, With a power supply of 5 V.
\begin{figure}[t!] 
\centering
\includegraphics[width=3in]{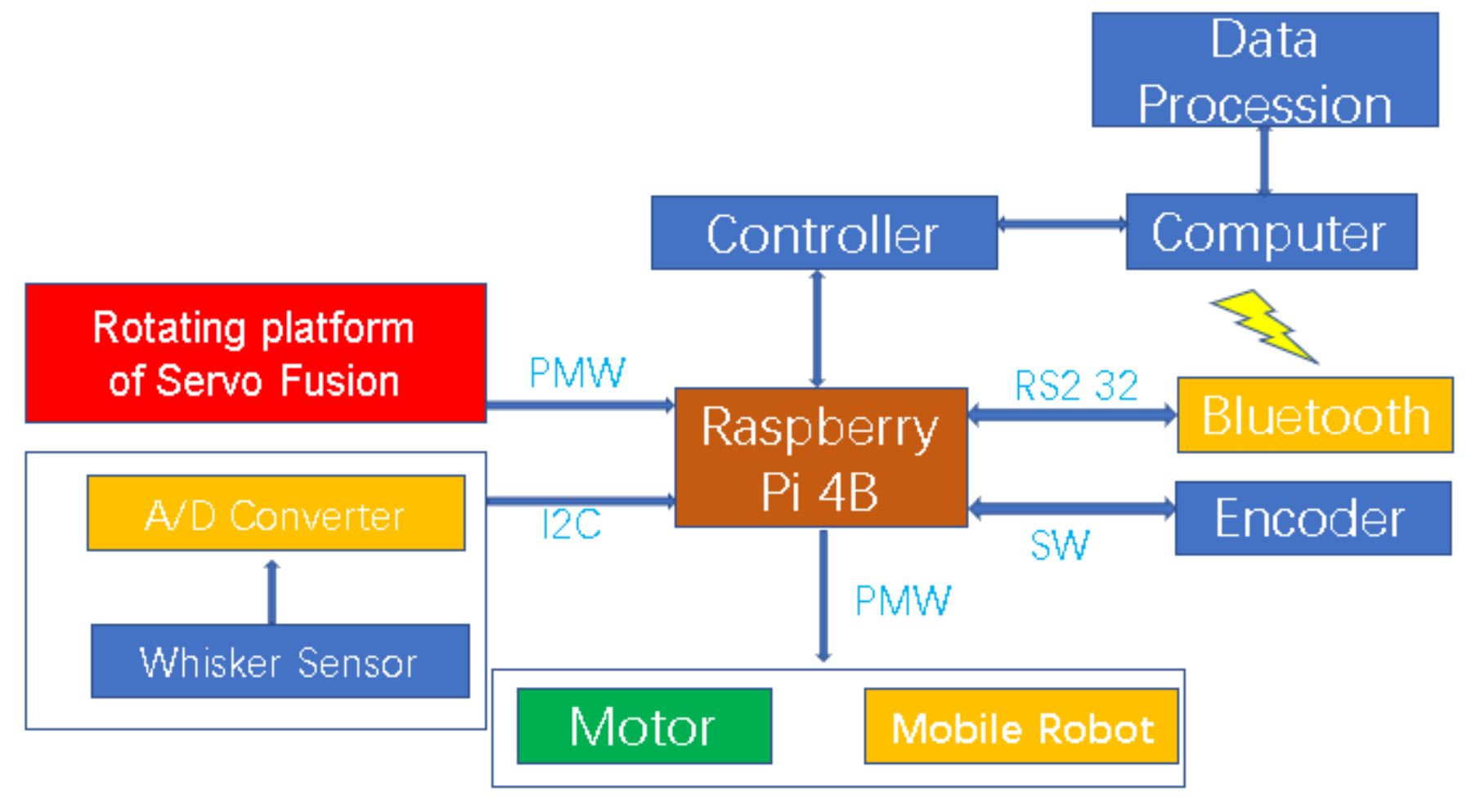}
\caption{Schematic diagram of the robot whisker system flow framework}
\label{Schematic diagram of the system flow framework}
\end{figure}

The robot system consists of a  whiskered tactile sensor  orthogonal  fixed in a 4-DOF servo fusion rotating platform, controlled by a Raspberry Pi 4 B.The whiskered tactile sensor has its own amplifier, and the raw voltage signals are sent to an ADS1115 analog to digital converter.

The  AD converters, servo motor driver as well as the 4-DOF rotating platform are connected to a Raspberry Pi 4B. The sensor data collection, robot's motion and rotating platform's movement are synchronized controlled by the Raspberry Pi 4B . 
All data are recorded in the raspberry pi on-board flash memory, and then transferred to a computer (1.60 GHz, 8 GB RAM) via Bluetooth for processing. Fig.\ref{Schematic diagram of the system flow framework} shows the system flow framework.

\begin{figure*}[t!]
\begin{minipage}[c]{\textwidth}
\small
\begin{eqnarray}
\label{eq:y}
y(x,t) &=& \sum_{i=1}^5 (- \frac{ \mathcal{A} \mathcal{B} \mathcal{C} }{ \mathcal{D} }), \\ \nonumber
\mathcal{A} &=& 2\,a\,h_{b}\,l^4\,{\omega _{b}}^2\,\rho \,{\mathrm{e}}^{-\frac{{D_{i}}^2\,t\,\zeta \,\sqrt{\frac{E\,I}{a\,\rho }}}{l^2}}\,\sin\left(\omega _{b}\,t\right)\,\left(\cos\left(D_{i}\right)-1\right)\,\left(\mathrm{cosh}\left(D_{i}\right)-1\right)\,\left(\cos\left(D_{i}\right)+\mathrm{cosh}\left(D_{i}\right)\right), \\ \nonumber
\mathcal{B} &=& \mathrm{sinh}\left(\frac{D_{i}\,x}{l}\right)-\sin\left(\frac{D_{i}\,x}{l}\right)+\frac{\left(\cos\left(\frac{D_{i}\,x}{l}\right)-\mathrm{cosh}\left(\frac{D_{i}\,x}{l}\right)\right)\,\left(\sin\left(D_{i}\right)+\mathrm{sinh}\left(D_{i}\right)\right)}{\cos\left(D_{i}\right)+\mathrm{cosh}\left(D_{i}\right)}, \\ \nonumber
\mathcal{C} &=& \zeta \,\sin\left(\frac{{D_{i}}^2\,t\,\sqrt{1-{\zeta }^2}\,\sqrt{\frac{E\,I}{a\,\rho }}}{l^2}\right)-{\mathrm{e}}^{\frac{{D_{i}}^2\,t\,\zeta \,\sqrt{\frac{E\,I}{a\,\rho }}}{l^2}}\,\sqrt{1-{\zeta }^2}+\cos\left(\frac{{D_{i}}^2\,\sqrt{E}\,\sqrt{I}\,t\,\sqrt{1-{\zeta }^2}}{\sqrt{a}\,l^2\,\sqrt{\rho }}\right)\,\sqrt{1-{\zeta }^2}, \\ \nonumber
\mathcal{D} &=& {D_{i}}^4\,E\,I\,\sqrt{1-{\zeta }^2}\, (3\,\mathrm{sinh}\left(D_{i}\right)\,{\cos\left(D_{i}\right)}^2\,\mathrm{cosh}\left(D_{i}\right)-D_{i}\,{\cos\left(D_{i}\right)}^2-3\,\sin\left(D_{i}\right)\,\cos\left(D_{i}\right)\,{\mathrm{cosh}\left(D_{i}\right)}^2+ \\ \nonumber
& & 3\,\mathrm{sinh}\left(D_{i}\right)\,\cos\left(D_{i}\right)+D_{i}\,{\mathrm{cosh}\left(D_{i}\right)}^2-3\,\sin\left(D_{i}\right)\,\mathrm{cosh}\left(D_{i}\right)+2\,D_{i}\,\sin\left(D_{i}\right)\,\mathrm{sinh}\left(D_{i}\right) ).
\end{eqnarray}
\end{minipage}
\end{figure*}

\section{Preliminary Theoretical Study}
\label{theoreticalstudy}
The sensor vibration characteristics are studied as a proof-of-concept for the idea presented.

\subsection{Cantilever Beam Vibration with Base Excitation}

The sensor is modelled as an equivalent cantilever beam with uniform mass under base excitation (Fig. \ref{Hall Sensor Model}).
The steady state response of such system for sinusoidal base excitation $y_b = h_b \sin(\omega_b t)$ can be found based on the derivations in \cite{tao_dynamic_2014} as a summation over the beam first five mode shapes in Eq. \ref{eq:y},
where $y(x,t)$ is the beam displacement at any length $x$ and time $t$, $h_b$ and $\omega_b = 2 \pi f$ are the base excitation amplitude in [m] and frequency in [rad/s], $f$ is the base excitation frequency in [Hz], $D = [ 1.8751,~ 4.6941,~ 7.8548,~ 10.9955,~ 14.137]$ is a set of constants related to a cantilever beam mode shapes \cite{tao_dynamic_2014}, and $\zeta = 0.04$ is the modal damping based on stainless steel material.
The equivalent beam has a length $l$ similar to the coil spring axial length $l_c=l=60$ [mm] and cross-section area $a$ similar to the spring wire cross-section area $a_w=\pi r_w^2$, where $r_w=0.5$ [mm] is the wire radius.
The beam unit length density is $\rho = C \rho_w $ where $\rho_w = 8050$ [Kg/$m^3$] is the wire material (stainless steel) density, and $C = l_w / (n p)$ is a correction factor to account for the coil shape based on the spring wire pitch $p=l_s/n$, overall length $l_w = n \sqrt{(2 \pi r)^2 + p^2 }$, and spring number of coils $n=13$.
We may assume that the beam bending modulus $E I$ is equivalent to the spring wire torsional modulus $G_w J_w$, where $G_w=E_w/3=70$ [GPa] is the wire material (stainless steel) shear modulus, and $J_w=\pi r_w^4 / 4$ is the wire cross-section $2^{\rm nd}$ moment of area.

\subsection{Sensor Modal Analysis}

\begin{figure}[t!] 
\centering
\includegraphics[width=0.49\textwidth]{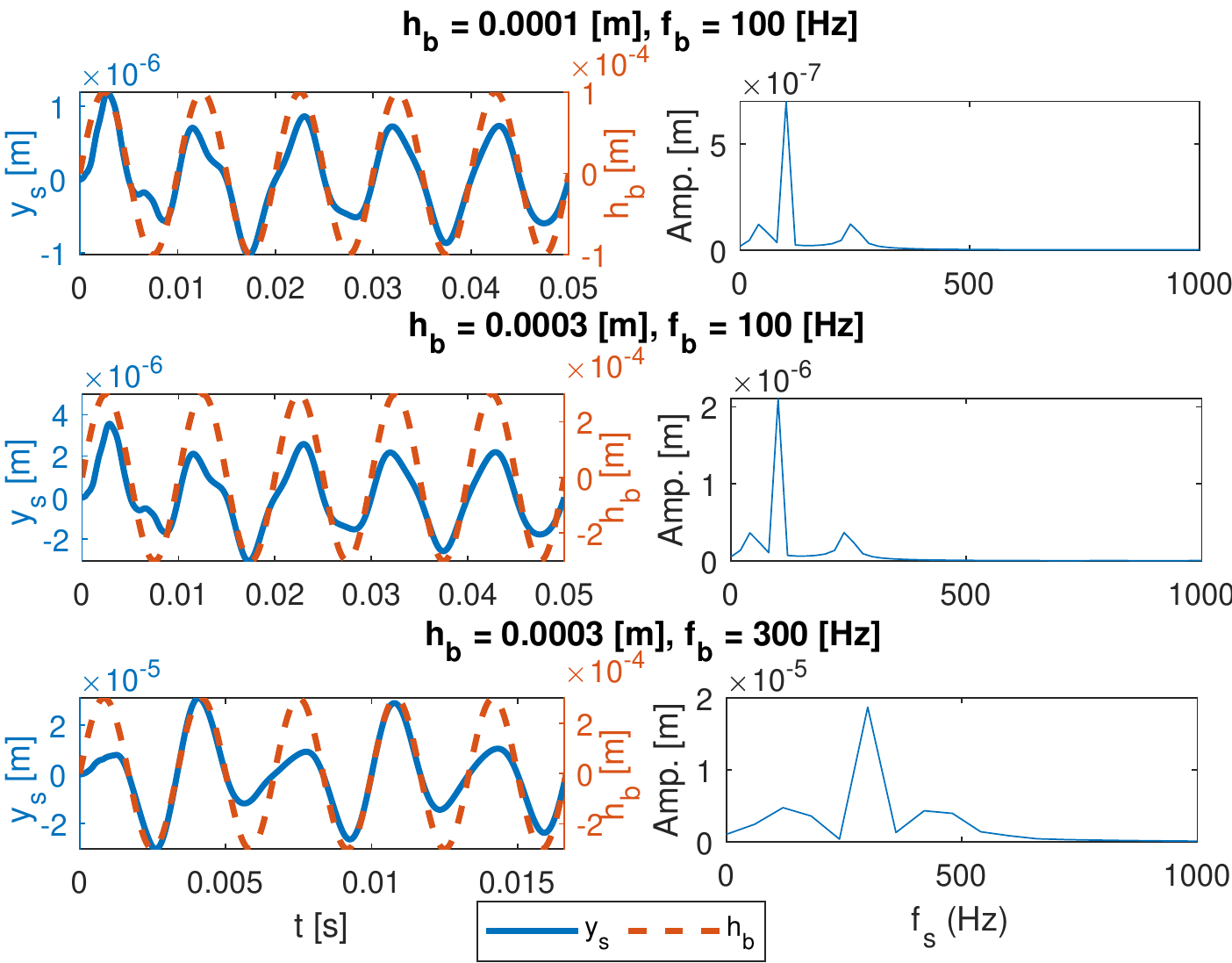}
\caption{Sensor displacement $y$ and FFT analysis due to various base excitation frequencies $f_b$ and amplitudes $h_b$.}
\label{fig:vibration}
\end{figure}

\begin{figure}[t!] 
\centering
\includegraphics[width=0.49\textwidth]{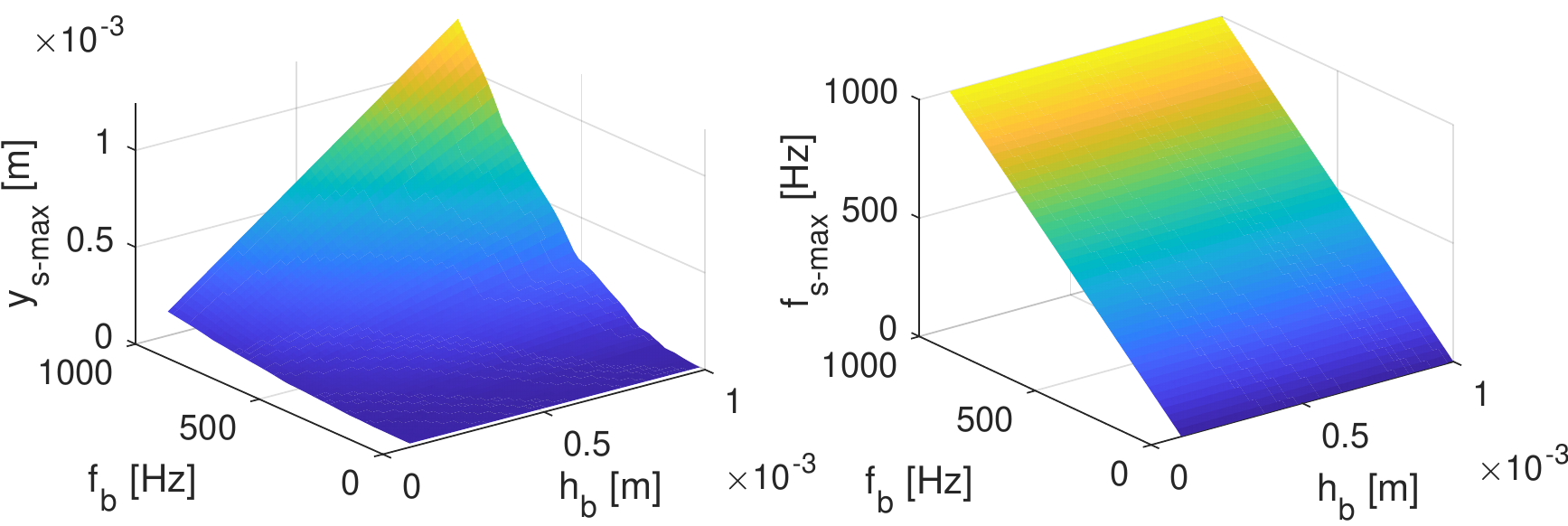}
\caption{Sensor displacement maximum value $y_{s-{\rm max}}$ and dominant modal frequency $f_{s-{\rm max}}$ vs. base excitation height $h_b$ and frequency $f_b$.}
\label{fig:surf}
\end{figure}

The simulation results for the sensor lateral displacement $y_s=y(x_s,t)$, where $x_s=5$ [mm] is the sensor distance from the beam base, and the signal FFT (Fast Fourier Transform) analysis in response to different excitation parameters ($h_b = 0.1 ~\& ~0.3$ [mm] and $f_b = 100 \& 300$ [Hz]) are plotted in Fig. \ref{fig:vibration}.
Distinctively different signal profiles and dominant frequency are observed for different excitation frequencies (i.e. due to ground texture).
The signal dominant frequency is the same as the base excitation frequency.

Fig. \ref{fig:surf} shows the sensor maximum displacement $y_{s-{\rm max}}$ in [m] and dominant modal frequency $f_{s-{\rm max}}$ in [Hz] for different excitation frequency $f_b$ and amplitude $h_b$.
It is observed that $f_{s-{\rm max}}$ is similar to $f_b$ and does not vary with $h_b$.
This shows that it is possible to successfully classify the excitation frequency $f_b$ (i.e. ground profile texture) based on the sensor signal dominant frequency $f_{s-{\rm max}}$.
$y_{s-{\rm max}}$ is mostly driven by $h_b$  for small excitation frequencies $f_b$, but affected by both the $f_b$ and $h_b$ for higher values of $f_b$.
As a result, the sensor signal amplitude $y_{s-{\rm max}}$ is not solely enough to determine valid information about the base excitation.
However, by classifying the excitation frequency $f_b$ based on the sensor signal modal analysis, the sensor signal amplitude $y_{s-{\rm max}}$ has enough information to determine the excitation amplitude $h_b$ (i.e. ground profile height).
Sudden changes in the slope of $y_{s-{\rm max}}$ vs. $f_b$ shows that sharper changes in the sensor signal should be anticipated as a result of profile texture variations for a higher frequency base excitation (e.g. rough train).

The above analysis indicates that the sensor signal can provide enough information for classifying the sensor base excitation (ground profile) if an appropriate method is employed to effectively handle the real-world uncertainties.
Such a method is discussed later in this paper.

\section{Data Processing and  deep Neural Network Training}
\label{dnn}
 The entire terrain surface identification and  recognition process are divided into three phases: 1) whisker-based off-line data collection and processing; 2) off-line training based on machine learning; 3) whisker-based online surface classification and recognition.
The whisker should be designed properly and the classification models need to  be trained off-line accurately, in order to achieve a higher success rate for identifying similar terrain surfaces. 
\begin{figure}[t!] 
\centering
\includegraphics[width=3.5in]{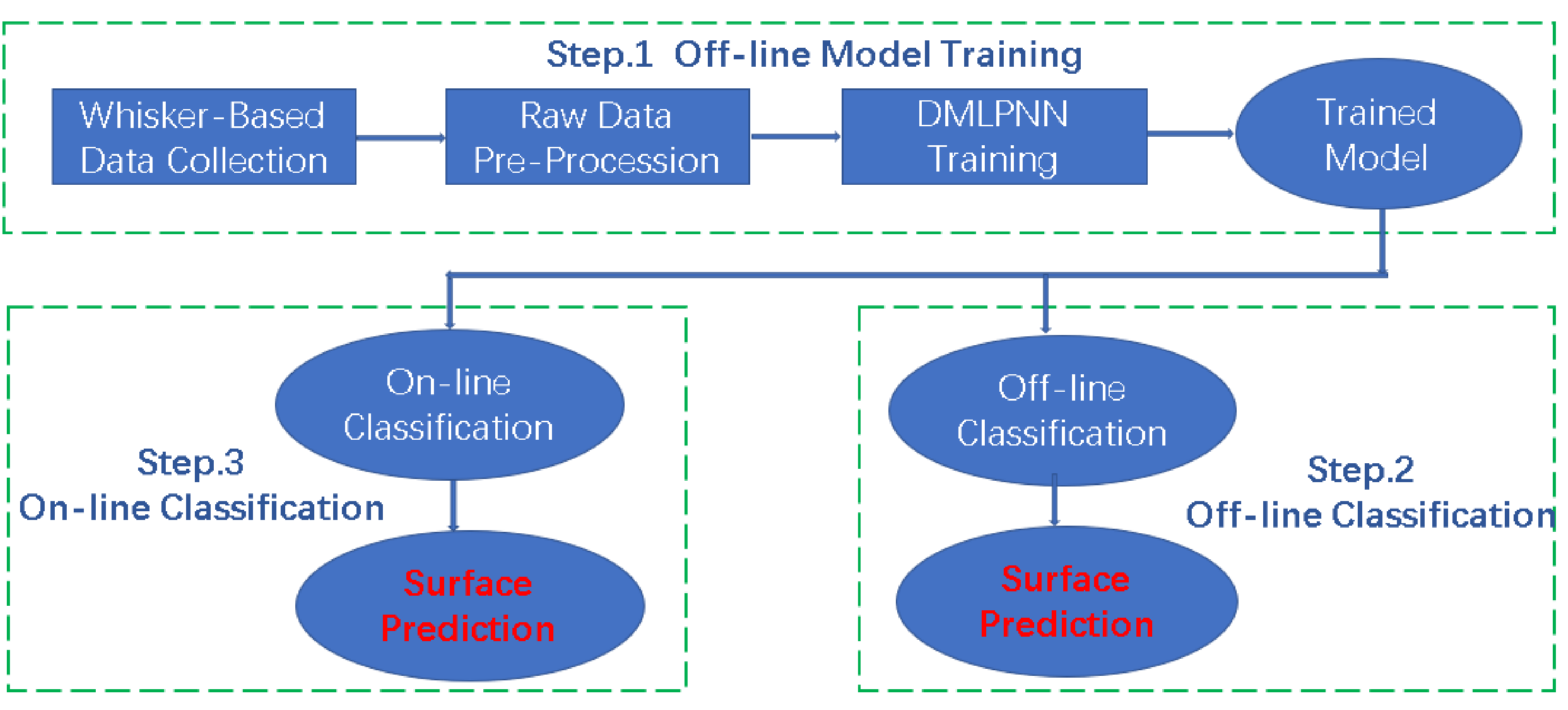}
\caption{Diagram of whisker vibration data collection and procession}
\label{Schematic diagram of data collection and procession}
\end{figure}

\subsection{Data Feature Extraction  }
The whiskered robot needs to collect enough whiskered tactile sensor vibration information for model training by traversing different terrain surfaces several times.  
The   operating frequency of whiskered tactile sensor during experiments is  200 Hz, and the  tactile vibration data which is collected by the whiskered  sensor was pre-processed and then segmented for accelerating the model training speed.
Every segmented vector corresponds to one-second of  data from the whiskered tactile sensor , so a  $1 *200$ labeled surface vector ${\bf{x}}_i^{1*200}$ can be created:
 \begin{equation}
{\bf{x}}_i^{1*200} = [{v_1},{v_2},.....{v_{200}}] \to {S_i}\left| {_{i = 1,2....7}} \right.
 \label{qqq16}
\end{equation}
Where ${S_i}$ is the different terrains surfaces and ${{i = 1,2,3,4,5,6,7}}$ corresponds to flat, cement, brick , carpet, soft-grass, sand and asphalt terrain surface respectively. The raw vibration data is subsequently converted from the time domain to frequency domain .
The first stage of data procession is standardized, and every vibration vector unit is normalized
to a unit vector whose standard deviation is 1 and the mean value is 0.
 
Then,  the tactile vibration vector unit ${\bf{x}}_i^{1*200}$ was transformed from the time-domain to  the frequency domain ${\bf{x}}_{_{(i,f)}}^{1*200}$ through a  Fast Fourier Transformation. The Fast Fourier transformation  has been proven efficient and capable of classifying the difference between multiple terrains in Section III and  it significantly enhances the   Fourier transform speed when compared to other Fourier transform methods.

\subsection{Network Architecture }
\begin{figure}[t!] 
\centering
\includegraphics[width=3.5in]{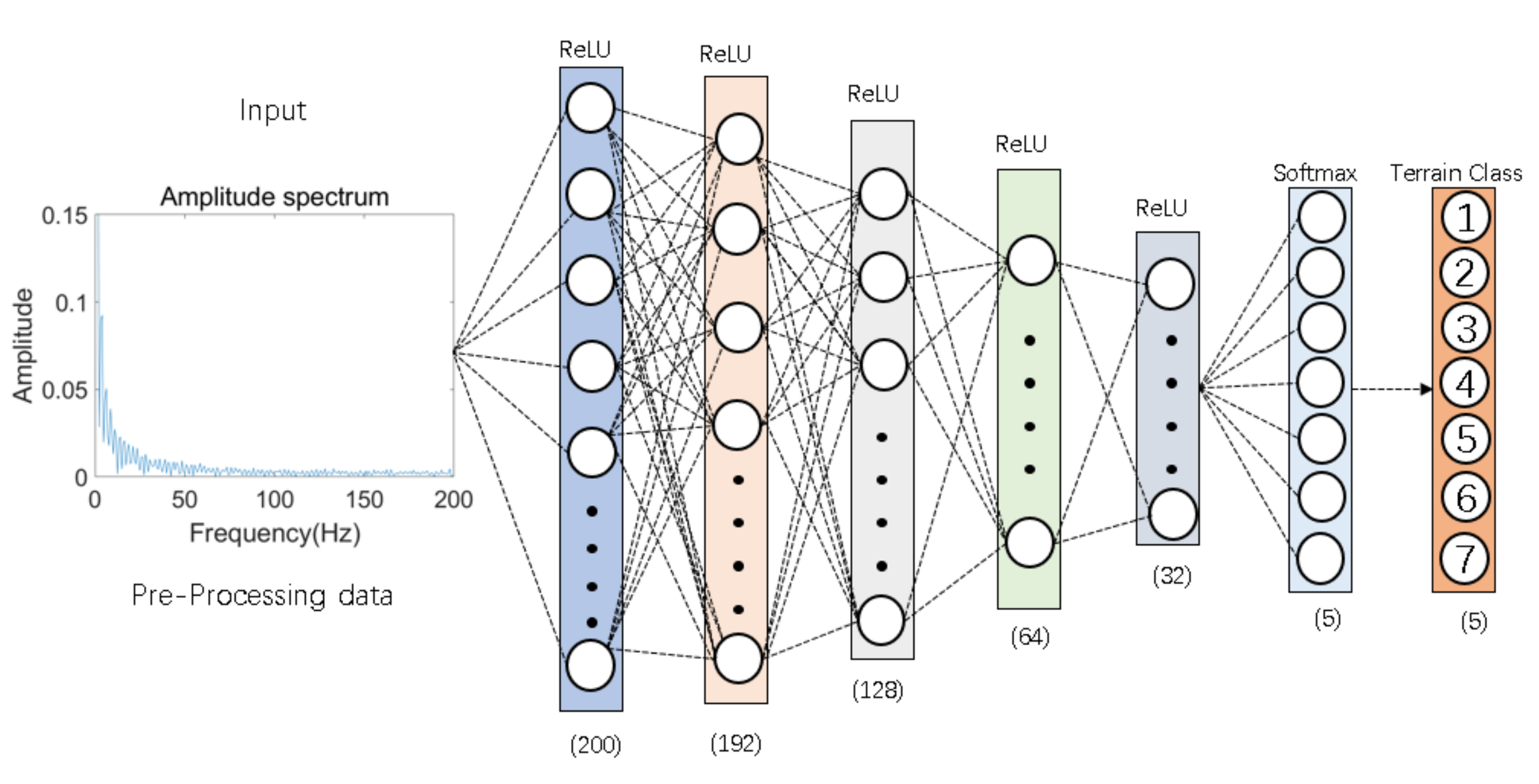}
\caption{Overview of Deep Multi-Layer  Perceptron Neural Network Pipeline.}
\label{Courant_2}
\end{figure}
As shown in Figure \ref{Courant_2}, a , deep neural network with seven-layers based on multi-layer perceptron is built to achieve the  different terrain surfaces identification. The activation functions of the first five layers are Rectified Linear Units, and the Soft-max function is introduced to better identify surface types. The cross-entropy loss function activation can be applied to evaluate the deviation matrix between the predicted and actual value.

For  different terrains, this project collects 5 minutes of tactile vibration data ${{\bf{X}}_{{\rm{raw}}}}$ for model training and terrain surfaces classification. These data are pre-processed and divided into 300 terrain feature vector unit ${\bf{x}}_{_{(i,f)}}$ based on above data feature extraction method. All the data are labeled and connected as a terrain feature class vector ${\bf{X}}_{_{(i,f)}}$.

During the training and classification period, 75 \% data  ${{\bf{X}}_{{\rm{train}}}}$ are randomly selected from the ${\bf{X}}_{_{f}}$for model training, and the remaining 25\%  data ${{\bf{X}}_{{\rm{test}}}}$ are used to test the performance of network. 
The input layer of this network consists of 200 neurons which corresponds to
the dimension of  the surfaces feature vector unit ${\bf{x}}_{_{(i,f)}}^{1*200}$, and the output layer includes 7 neurons, corresponding to the seven different terrain surface types considered here. 
All layers of the network are fully connected. After the training is completed, this network  is used to classify the remaining  25\% testing vectors   data ${{\bf{X}}_{{\rm{test}}}}$  and it returns an estimation of the terrain type.
By trial and error, all the network’s training parameters were updated, but it is impossible to make it optimal.

\section{RESULTS}
\label{results}
To verify the reliability and robustness of the aforementioned whiskered tactile sensors and sensor vibration characteristics analysis based on reservoir
computing, surface identification experiments were conducted on seven different terrains at the same speed.
Then, the influence of the  mobile robot running speed on the whiskered tactile sensor’s identification capability  has also been analyzed and discussed from nonlinear dynamic  resonance perspectives based on the theory presented in Section III.

\subsection{Terrain Surfaces Identification }

\begin{figure}[t!] 
\centering
\includegraphics[width=3.5in]{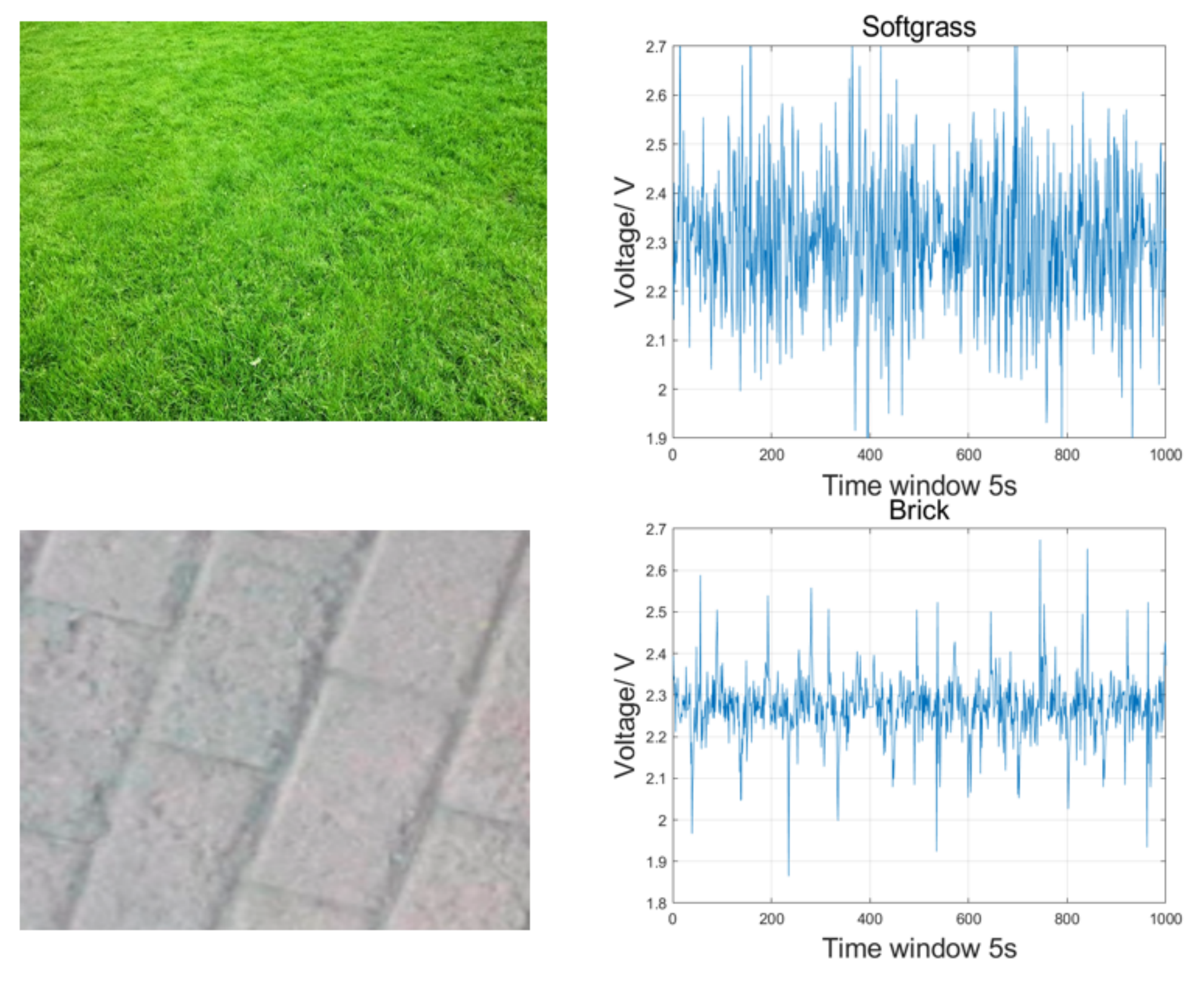}
\caption{Comparison of two example terrain surface(top two:soft-grass;the bottom two:brick) and corresponding raw vibration data. The time window of this vibration voltage data is 5 seconds }
\label{Example of terrain Surface}
\end{figure}

Based on the above methods, the whiskered robot was controlled by a Raspberry Pi 4 system to move on 7 different terrain surfaces: brick, cement, flat terrain, carpet, soft-grass, sand and a soft-soil surface respectively.
All raw collected-data are pre-processed and   the deep   multi-layer   perceptron   neural   network is pre-trained offline properly. Part of the sample experiment terrain surface and its corresponding vibration data are given in Figure \ref{Example of terrain Surface}.

This experiment collects 5 minutes of vibration data for each terrain surface.
By randomly extracting 75 \% data ${{\bf{X}}_{{\rm{train}}}}$ from the terrain tactile vibration database as the training data, and  the rest 25\% of data ${{\bf{X}}_{{\rm{test}}}}$ were used to test the performance of this network. The experimental parameters are set as shown in Table \ref{Experiments Parameter Setting}.
In all, 20 random experiments were conducted to avoid accidental errors caused by random trials.
\begin{table}[t]
\centering
\caption{
Experiments Parameter Setting - 300 sets (Training data: 75\%; Test Data 25\% ) }
\begin{tabular}{ccc} 
	\hline
  Parameters   & Value    \\ 
	\hline
Robot Velocity & 0.2 m/s &  \\
Sampling Frequency  & 200 Hz &  \\
Time-window  & 1 sec &  \\

	\hline
\end{tabular}
\label{Experiments Parameter Setting}
\end{table}


\begin{figure}[t!] 
\centering
\includegraphics[width=3.5in]{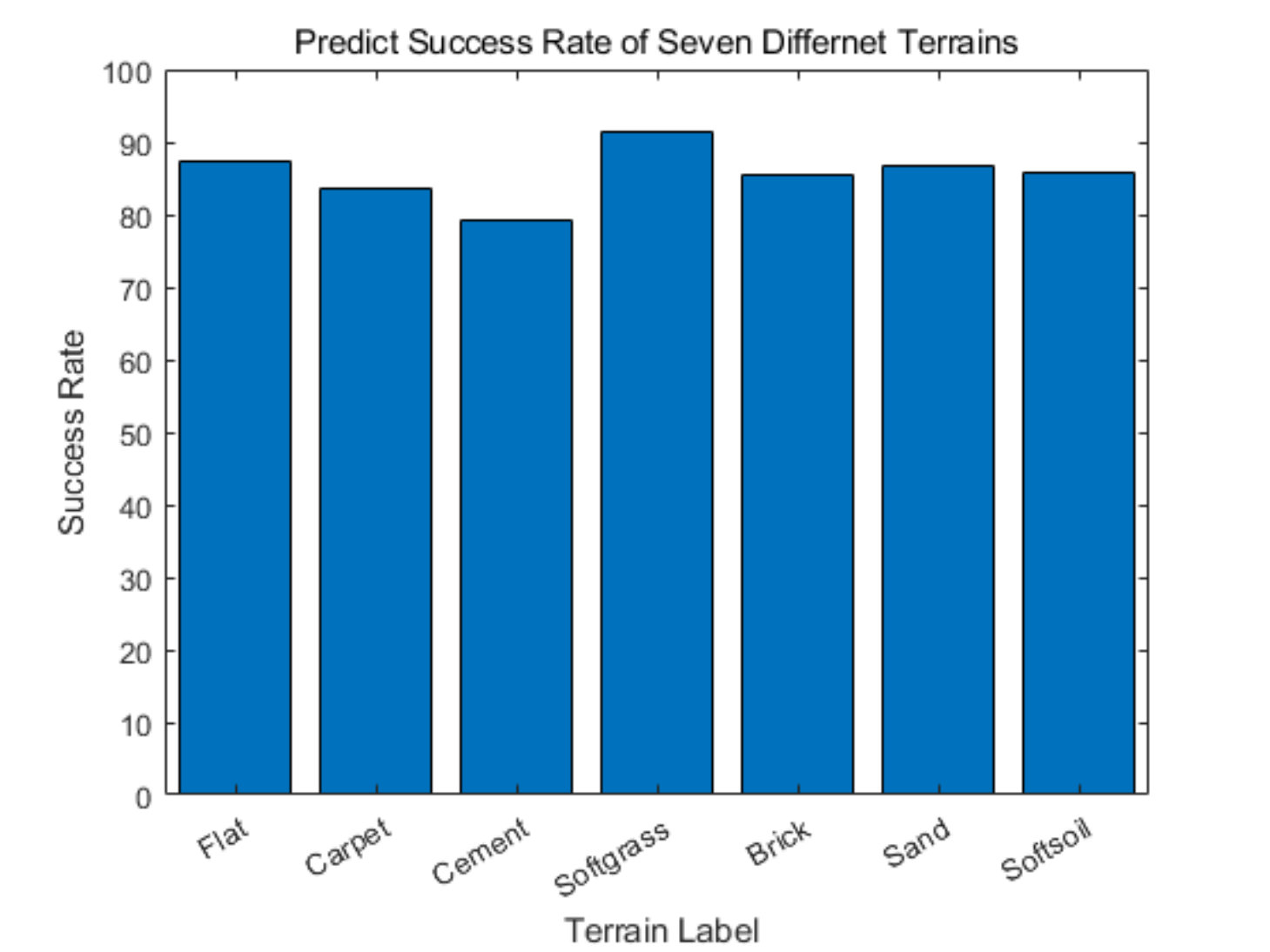}
\caption{Prediction rate of seven terrains surface}
\label{prediction success rate of seven terrains surface 2222}
\end{figure}

As can be seen from the Figure \ref{prediction success rate of seven terrains surface 2222}, the predicted classification success rates for flat terrain, carpet, cement, soft-grass, brick, sand and softsoil terrain surface are $87.3\%$, $83.6\%$, $79.2\%$, $91.3\%$, $85.4\%$, 86.6\%, and 85.8\% respectively. The experimental results indicate that the whiskered tactile sensor and the deep multi-layer perception neural network have good recognition and identification capabilities of different terrains, with an average  success rate of about 85.6\%.The results shows that the steady state response  of  nonlinear  vibration  dynamics  of whisker sensor can  be  used  to classify different terrains. 

\begin{figure}[t!] 
\centering
\includegraphics[width=3.5in]{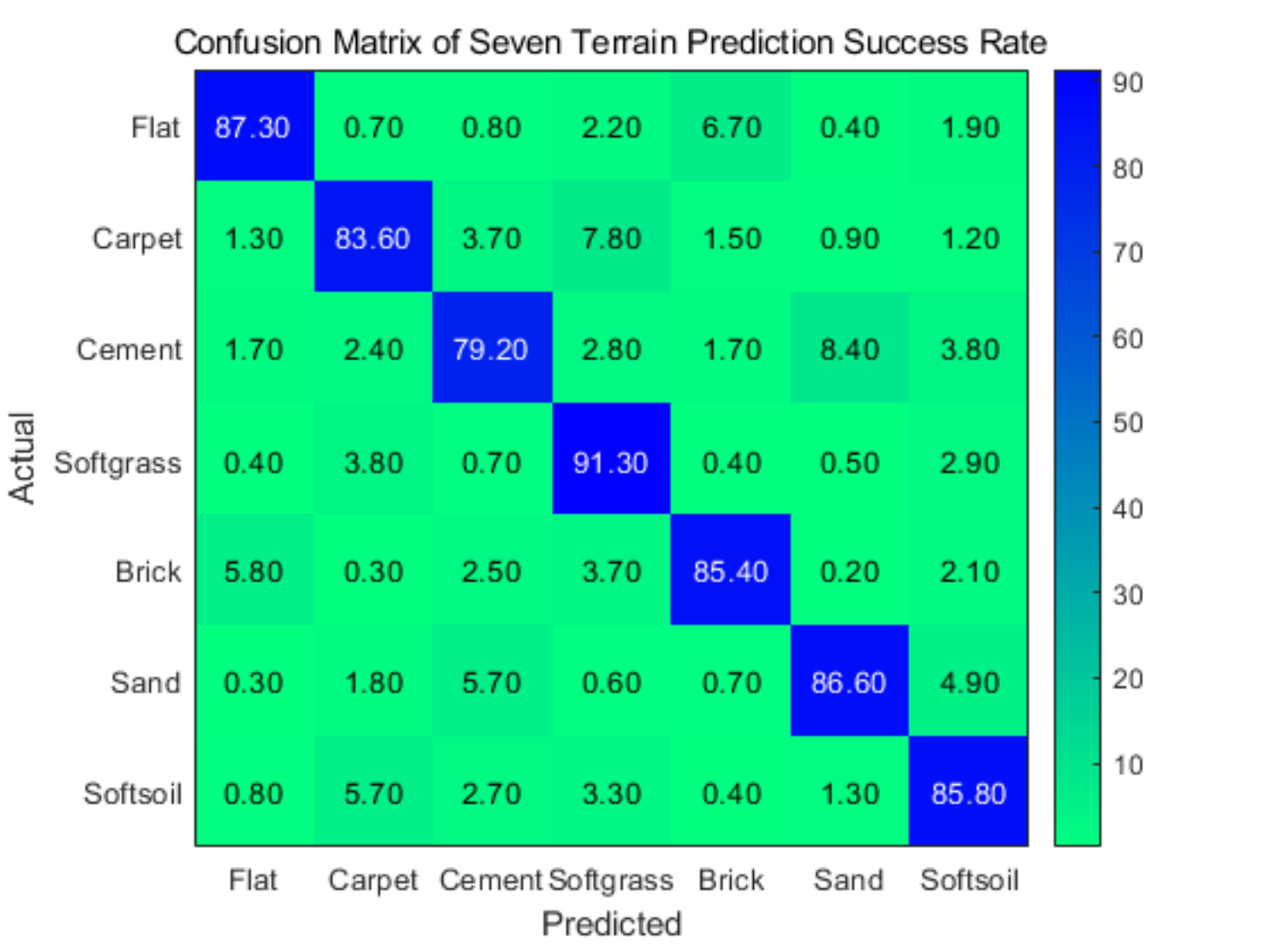}
\caption{Confusion matrix of seven terrain prediction success rate}
\label{Confusion matrix111}
\end{figure}

Based on the confusion matrix in Figure \ref{Confusion matrix111}, this network has high identification and classification success rate for flat and soft grass terrain surface because they have distinct vibration features based on their vibration signals. However, there is a high confusion rate between the flat surface and the brick surface which can be explained by the similar gap between the two adjacent units. In the case of the cement terrain, this could have a complicated surface including features such as gaps,  a smooth or rugged surface which make the network more challenging to classify it from others.

Combined  with  the  simulation  results  in  Section III, the results indicate that robot’s base excitation vibration
amplitude (i.e. ground profile height) affects the sensitivity of whisker  sensor , and the sensor signal can provide enough information for classifying the sensor base
excitation (terrain profile) by employing deep multi-layer
perceptron neural network.

\subsection{Different Speed Experiments}

The result of the previous section indicates that the whiskered tactile sensor and the  proposed deep multi-layer perceptron neural  network reported in this paper has good
classification and identification capabilities  on seven terrains ${S_i}$ at a constant velocity 0.2 m/s.
The influence of different speeds on the sensor and model prediction accuracy has been analyzed.
Data collection was performed at speeds of $0.1$ m/s, $0.15 $m/s, $0.2 $m/s, $0.25$ m/s and $0.3$ m/s respectively. All parameters during this experiment are the same as in the previous section. The identification accuracy results are shown in Table \ref{Comparisontable11111111111}.
\begin{table}[t]
\centering
\tiny
\caption{
Identification Accuracy of Seven Terrain Surfaces at different speeds}
\begin{tabular}{cccccccc}
	\hline
Speed & ${S_1}$   & ${S_2}$  &${S_3}$  &  ${S_4}$& ${S_5}$  & ${S_6}$& ${S_7}$  \\ 
	\hline
$0.10m$/s & $87.6\%$ & $78.8\%$&$82.3\% $& $88\%$&$84.8\%$& $86.9\% $& $74.8\%  $\\
$0.15m/s $ & $86\%$ & $85.2\%$&$88.6\%$ & $83.2\%$&$82\%$& $93.2\% $& $82\% $ \\
$0.20m/s $& $87.3\%$ & $83.6\%$&$79.2\% $& $91.3\%$&$85.4\%$&$ 86.6\% $& $85.8\% $ \\
$0.25m/s$  & $90.6\%$ & $93\%$&$90.3\%$ & $81.3\%$&$89.1\%$& $79.3\% $& $91\% $ \\
0.30m/s &92.5\% & $78.5\%$& $82.2\% $& $85.5\%$ & 92.5\%& 84.2\% & 78.5\% \\
\hline
\end{tabular}
\label{Comparisontable11111111111}
\end{table}

\begin{figure}[t!] 
\centering
\includegraphics[width=3.5in]{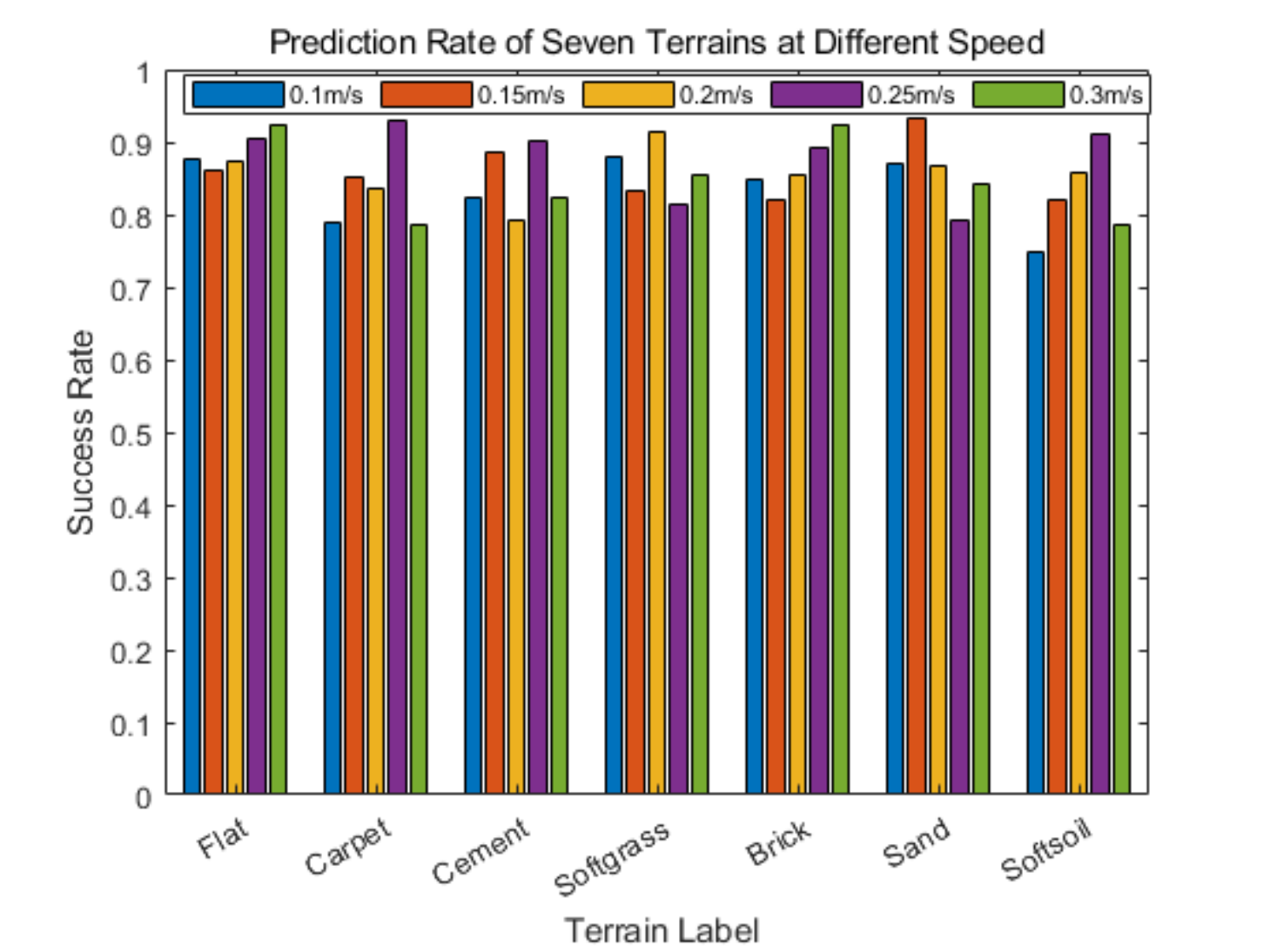}
\caption{Identification Accuracy of Seven Terrain Surfaces at different speeds.The colour of blue, orange,yellow, purple and green represent the speed of 0.1 m/s, 0.15 m/s, 0.2 m/s, 0.25 m/s and 0.3 m/s respectively. }
\label{prediction rate of seven terrains surface 234342}
\end{figure}

\begin{figure}[ht!] 
\centering
\includegraphics[width=3in,height=2in]{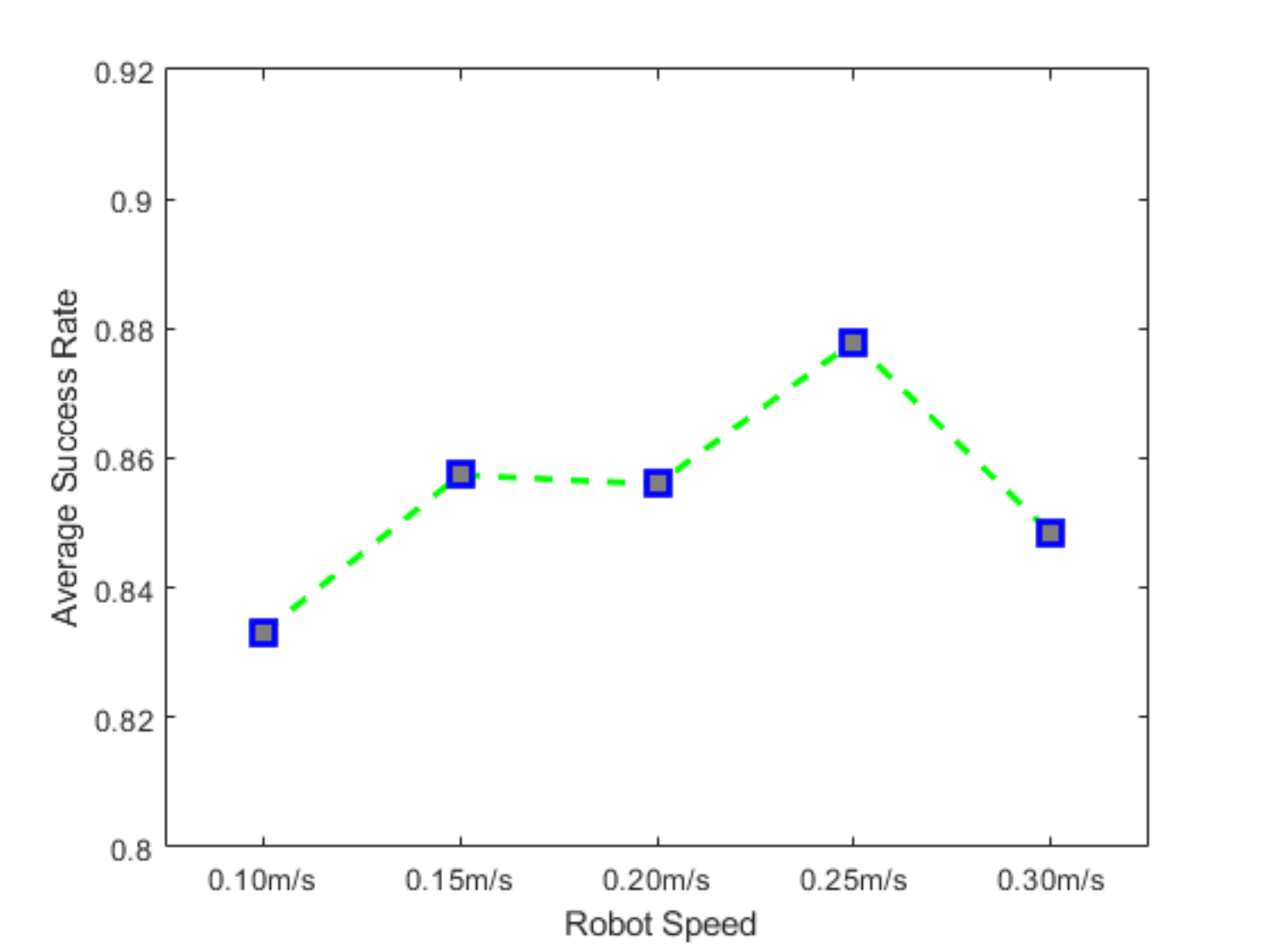}
\caption{Average prediction accuracy of seven terrains at different speeds}
\label{prediction rate of seven terrains surface 23432}
\end{figure}

Looking at Figure \ref{prediction rate of seven terrains surface 234342} and Figure \ref{prediction rate of seven terrains surface 23432},   it can be seen that this system has good identification accuracy for seven terrains surface at different speeds. The average prediction accuracy of the seven terrains  at different speeds are: 83.31\%, 85,74\%, 85.60\%, 87.8\%, and 84.84\%.
Therefore, speed has
no consistent effect on identification success rate. However,
there are still some details here that can be used as a
reference to improve the identification accuracy.
Firstly, the average identification accuracy at a speed of v = 0.25 m/s is higher than the other four operating states reaching 87.8\%, which indicates that the speed of mobile robot has  influence on identification accuracy. This may be caused by the dynamic resonance of the nonlinear experiment system. The spring beam and the robots car  achieve a better resonance effect at $0.25$m/s, which makes the spring vibration more obvious. Since the sensitivity of the sensor depends on the vibration of the spring, it has a better classification effect at this speed. This also echoes the simulation results of the Section \ref{theoreticalstudy}, indicating that robot’s base excitation vibration frequency affects the whisker tactile sensor sensitivity. Moreover, a nonlinear system has multiple resonance frequencies, this can also explain why good results are also achieved at the speed 0.15m/s but relatively poor at 0.2m/s.

Secondly, the accuracy of distinguishing between the brick and flat terrain surfaces increases when the robot speed increases.  Combined with modal analysis in  Section \ref{theoreticalstudy}, these results indicate that terrain surface profile amplitude and robots vibration frequency affects the sensor sensitivity in the same way. But, the surface amplitude is the dominant factor in affecting the sensor displacement for low frequency profile.

\section{CONCLUSIONS}
\label{conclusions}

Whiskers, vestibular system, and the cochlear are three examples of compliant mechanical systems in biological counterparts that solve realtime perception problems using nonlinear vibration dynamics. Work done on such reservoir computing systems show that even periodic sinusoidal external perturbations can lead to complex steady state dynamics in the compliant mechanical system including period bifurcation and frequency separation at local sites. This paper shows for the first time that the steady state response of nonlinear vibration dynamics can be used to classify textures even on flat terrain. We also show that a mobile robot can use speed control to move the perturbation frequencies to elicit unique frequency domain responses in a whisker to help terrain classification. Experiment results show that the novel whiskered sensor and the deep multi-layer perceptron neural network have good recognition and identification capabilities of different terrain surfaces for mobile robots.

In  the  future,  it will be interesting to investigate how realtime stiffness control of the whisker can be used as a control parameter to elicit steady state vibration frequency components unique to the texture and low frequency geometric features of a given terrain.

\addtolength{\textheight}{-12cm}   

\bibliographystyle{IEEEtran}
\bibliography{bibs}

\end{document}